%% file: main.tex
\documentclass[10pt,twocolumn,letterpaper]{article}

\usepackage{cvpr}              
\newcommand{\PIPS}{\texttt{POSTURE}}
\newcommand{\PIPSSF}{\texttt{SF-POSTURE}}

\usepackage{bbm}
\DeclareMathOperator*{\argmax}{arg\,max}
\usepackage{multirow}
\usepackage{booktabs}
\usepackage{amssymb}
\usepackage{pifont}
\usepackage{wasysym}
\usepackage{adjustbox}
\usepackage{microtype}

\usepackage{stmaryrd}
\usepackage{mathtools}

\newcommand{\xmark}{\ding{55}}%
\newcommand{\cmark}{\ding{51}}%

\usepackage[pagebackref,breaklinks,colorlinks,citecolor=blue]{hyperref}

\title{POSTURE: Pose Guided Unsupervised Domain Adaptation for Human Body Part Segmentation}

\author{Arindam Dutta$^{1,* }$ \ Rohit Lal$^{1, 2, * }$ \ Yash Garg$^{1 }$ \ Calvin-Khang Ta$^{1 }$ \ Dripta S. Raychaudhuri$^{1,3}$  \\ \ Hannah Dela Cruz$^{1}$ \ Amit K. Roy-Chowdhury$^{1 }$\\
$^{1}$University of California, Riverside $ \quad \quad ^{2}$ NASA-Impact $ \quad \quad ^{3}$AWS AI Labs \\
{\tt\small \{adutt020@, rlal011@, ygarg002@, cta003@, drayc001@, hdela004@, amitrc@ece.\}ucr.edu} 
}
\usepackage{comment}

\begin{document} 
\maketitle 
\input{sections/0_abstract}
\input{sections/1_intro}
\input{sections/2_related}

\input{sections/updated_method}

\input{sections/updated_experiments}

\input{sections/5_conclusion}

{
    \small
    \bibliographystyle{ieeenat_fullname}
    \bibliography{ref}
}


\end{document}

%% file: sections/0_abstract.tex
\textit{This work has been submitted to the IEEE for possible publication. Copyright may be transferred without notice, after which this version may no longer be accessible.} \\

\begin{abstract}
Existing algorithms for human body part segmentation have shown promising results on challenging datasets, primarily relying on end-to-end supervision. However, these algorithms exhibit severe performance drops in the face of domain shifts, leading to inaccurate segmentation masks. To tackle this issue, we introduce \PIPS: \underline{Po}se Guided Un\underline{s}upervised Domain Adap\underline{t}ation for H\underline{u}man Body Pa\underline{r}t S\underline{e}gmentation - an innovative pseudo-labelling approach designed to improve segmentation performance on the unlabeled target data. Distinct from conventional domain adaptive methods for general semantic segmentation, \PIPS~ stands out by considering the underlying structure of the human body and uses anatomical guidance from pose keypoints to drive the adaptation process. This strong inductive prior translates to impressive performance improvements, averaging 8\% over existing state-of-the-art domain adaptive semantic segmentation methods across three benchmark datasets. Furthermore, the inherent flexibility of our proposed approach facilitates seamless extension to source-free settings (\PIPSSF~), effectively mitigating potential privacy and computational concerns, with negligible drop in performance.

\end{abstract}

%% file: sections/1_intro.tex
\section{Introduction}
\label{sec:intro}

\begin{figure}[t]
    \centering
    \includegraphics[width =0.9\linewidth]{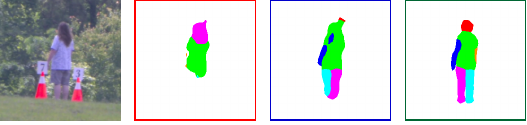}
    \caption{{\bf Need for domain adaptive human body part segmentation.} {\textit{Left to right}}: RGB image \cite{cornett2023expanding}, masks predicted from a model trained on synthetic data \cite{varol2017learning}, predictions from a SOTA UDA algorithm for semantic segmentation  \cite{vu2019advent}, and predictions from \PIPS. The model trained on synthetic data produces highly inaccurate predictions due to the inherent distribution shift between synthetic and real images. While domain adaptive semantic segmentation \cite{vu2019advent} improves the result, their inability to account for the inherent anatomical structure leads to sub-optimal segmentation. In contrast, \PIPS~ considers the underlying anatomical structure of the human body, delivering accurate segmentation masks for human body parts under domain shifts.}
    \vskip -4mm
    \label{fig:teaser_qual}
\end{figure}

Human body part segmentation, also known as semantic human parsing, is a fundamental computer vision task that involves the identification and delineation of various body components. Its significance extends across a wide variety of applications, including person re-identification \cite{kalayeh2018human}, conditional human image generation \cite{wu2021image}, virtual clothing try-on \cite{zhao2021m3d}, dense pose estimation \cite{guler2018densepose}, 3D human mesh reconstruction \cite{varol2017learning}, human de-occlusion \cite{zhou2020occlusion, zhao2021prior}, gait recognition \cite{zheng2023parsing}, and medical imaging \cite{voss2023multi}. While state-of-the-art human parsing approaches \cite{li2020self, ruan2019devil} achieve impressive performance on multiple datasets \cite{gong2017look, liang2015human, chen2014detect}, their reliance on supervised training is a significant hindrance, as acquiring and annotating diverse real-world data for training is expensive and not scalable. Furthermore, these algorithms exhibit limited generalization ability to out-of-distribution samples. Even minor domain shifts, such as the presence of Gaussian blur, tend to result in sub-optimal performance, as documented in \cite{zhang2023exploring}. While synthetic data offers an easy remedy for the initial challenge of obtaining annotations \cite{varol2017learning}, the issue of limited generalization persists due to the inherent distribution shifts between synthetic and real images. Figure~\ref{fig:teaser_qual} qualitatively illustrates the sub-optimal performance of a model trained on synthetic data \cite{varol2017learning} when applied to a real image. In light of this issue, we focus on the task of \emph{unsupervised domain adaptation (UDA) for enhancing the performance of human body part segmentation models in real-world scenarios}.

\begin{figure}[t]
    \centering
    \includegraphics[width = 0.9\linewidth]{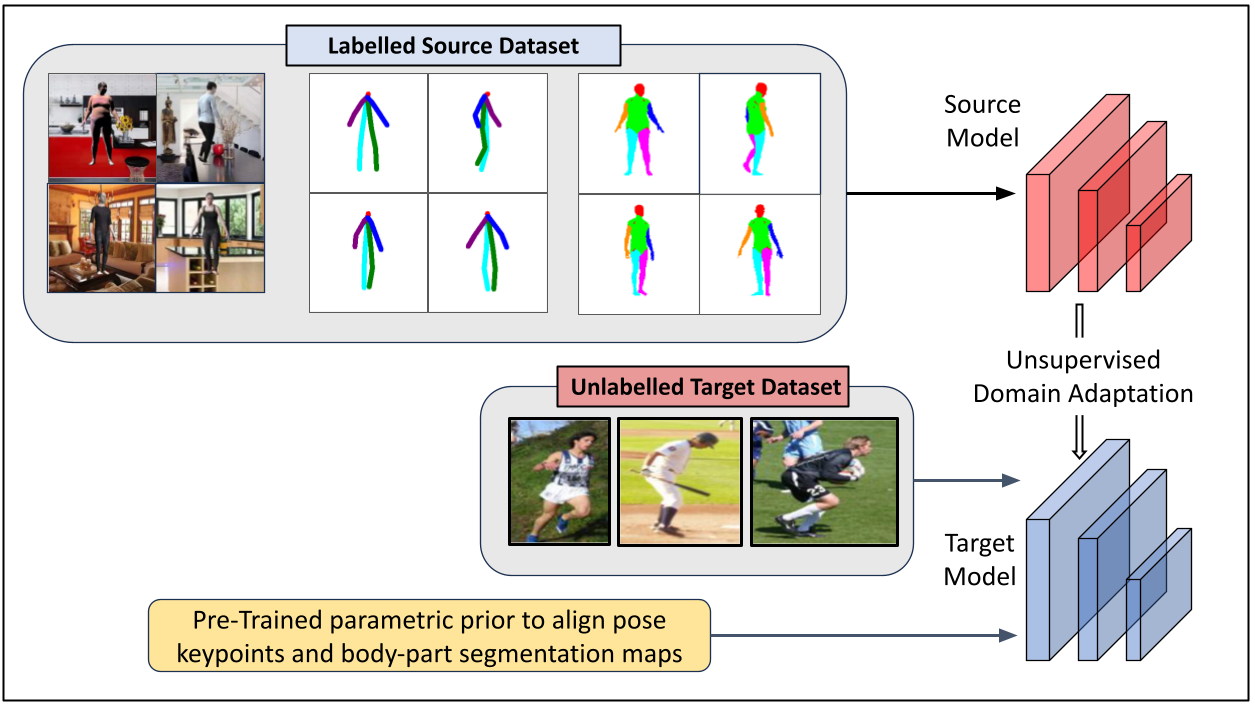}
    \caption{{\bf Problem overview.} We introduce \PIPS~ for domain-adaptive human body part segmentation. By leveraging confident pseudo-labels and explicit anatomical guidance from pose keypoints, \PIPS~ delivers improved segmentation masks in the presence of domain shifts. A pre-trained parametric mapping is utilized to align the estimated pose keypoints with body part segmentation maps. This mapping acts as strong prior in the face of domain shifts, disallowing the target model to overfit to anatomically implausible segmentation maps. We also extend \PIPS~ to source-free UDA settings (\PIPSSF), thereby addressing privacy and storage bottlenecks associated with using source data for adaptation.} 
    \label{fig:teaser_final}
    \vskip -4mm
\end{figure}

Numerous works have studied the problem of UDA in the context of semantic segmentation \cite{vu2019advent, paul2020domain, tranheden2021dacs}. Foundational works primarily utilized adversarial learning to bridge the domain gap, as discussed in \cite{tsai2018learning, vu2019advent, kim2020learning}. Another line of research focuses on self-training to improve pseudo-labels on the unlabeled target domain, using methods such as confidence regularization \cite{zou2019confidence} or consistency training \cite{ma2021coarse}. Augmentation-based self-training methods \cite{tranheden2021dacs, araslanov2021self, yang2020fda} have also shown promising results in this field. However, these algorithms primarily focus on urban scene segmentation rather than human body part segmentation and do not consider the underlying anatomy of the human body. Consequently, their performance in addressing our specific problem is sub-optimal.


To overcome the aforementioned bottlenecks, we propose \PIPS~ (\textit{\underline{Po}se Guided Un\underline{s}upervised Domain Adap\underline{t}ation for H\underline{u}man Body Pa\underline{r}t S\underline{e}gmentation}), a novel self-training UDA framework for body part segmentation that utilizes explicit anatomical information to drive the adaptation process. Specifically, \PIPS~ utilizes pose keypoints to inform the body part segmentation model of the underlying anatomy. This is done via a parametric mapping between the sparse pose keypoints and the dense body part segmentation maps that is learned offline \cite{duttapoise}. Using this mapping, we enforce a consistency regularization between the body part segmentation results predicted from the image and those predicted from the pose. The pose information allows us to automatically overcome incorrect predictions resulting from the distribution shifts. This is due to the fact that the mapping does not use any information from the RGB images, making it domain-agnostic, unlike the body part segmentation model. This pose-driven prior also allows us to extract reliable pseudo-labels \cite{karim2023c} from the body part segmentation model under adaptation, thereby avoiding learning trivial solutions. 

Despite its advantages, like all traditional UDA algorithms, \PIPS~ requires access to source data which may be problematic for privacy reasons. This is not a problem when the source data is synthetic, however, in situations where the source model is trained on a proprietary real-world dataset, it can be problematic to release this data for downstream adaptation due to privacy concerns. This is especially true for body part segmentation datasets since they involve human subjects \cite{xiang2022being, schwartz2011pii}. Moreover, employing source data for adaptation might not be feasible due to limitations in storage and computational resources. While source-free UDA approaches for urban scene segmentation have been proposed in \cite{liu2021source, huang2021model}, both of these methods depend on using the source data in indirect ways. The algorithm in 
\cite{liu2021source} regenerates the source data using an adversarial learning setup, while \cite{huang2021model} requires a model that was pre-adapted using source data, thereby violating privacy and computational benefits associated with source-free UDA.


To deal with the potential emergence of these issues, we also extend \PIPS~ to the source-free UDA setting \cite{liang2020we}, where it is termed as \PIPSSF. While the absence of source data in UDA settings often results in forgetting of the source knowledge \cite{liu2021ttt++}, the pre-trained parametric prior  in our framework serves as a robust regularizer, penalizing the model for anatomically implausible segmentation masks. This mitigates forgetting and assists \PIPSSF~in achieving results comparable to \PIPS. Note that our learned parametric mapping does not use RGB images for training, thereby preserving the privacy aspect of \PIPSSF. Figure~\ref{fig:teaser_final} gives an overview of our problem formulation. 

\noindent \textbf{Contributions.} In summary, we make the following key contributions: \begin{itemize}
    
    \item We address the problem of adapting a pre-trained human body part segmentation model to an unlabeled dataset in the presence or absence of source data.

    \item We introduce \PIPS, a simple and intuitive domain adaptation algorithm which addresses the problem at hand. \PIPS~ leverages confident pseudo-labels from the model under adaptation and pose-based anatomical information to drive the adaptation process. 
    
    \item Further, we show that \PIPS~ can be extended to source-free settings (\PIPSSF), thereby alleviating the privacy and computational bottlenecks of \PIPS.

    \item Extensive experimental validation on challenging domain adaptive scenarios establish \PIPS~ and \PIPSSF~as superior algorithms for human body part segmentation under domain shifts. 
\end{itemize} 


%% file: sections/2_related.tex
\section{Related Works}
\label{sec:related}

\noindent \textbf{Semantic Human Parsing.} Semantic human parsing, or human semantic segmentation, involves categorizing each pixel in an image of a person into distinct body parts or clothing, providing a detailed understanding of the human body in images. Pioneering works in this field include active template regression \cite{liang2015deep}, an adversarial learning algorithm for global and local consistency \cite{luo2018macro}, and a pose-parsing synergy based architecture \cite{liang2018look}. Authors of \cite{ruan2019devil} integrated contextual modules with edge information, while \cite{li2020self} introduced a self-correction algorithm which learns from noisy labels. Authors of \cite{zhang2020correlating} explored cross-task consistency between pose estimation and parsing. However, it is worth noting that these methods are entirely supervised learning algorithm, thus requiring costly per-pixel annotations. To address this, researchers have explored weakly-supervised methods, as demonstrated by Zhao {\it et. al.} \cite{zhao2022pose} using pose keypoints and Lin {\it et. al.} \cite{lin2020cross} in a domain adaptive scenario with available pose keypoints on the target dataset. Note that, our approach does not require access to ground-truth pose keypoints on the unlabeled target dataset whereas both \cite{zhao2022pose, lin2020cross} assumed availability of the same. 

\noindent \textbf{Human Pose Estimation.} 
Human pose estimation involves locating anatomical landmarks such as the head, elbows, and shoulders in 2D or 3D coordinates. State-of-the-art algorithms such as \cite{xiao2018simple, sun2019deep} excel on challenging academic datasets. However, these algorithms struggle to generalize to new images as these models are typically trained under supervised settings. Addressing this, \cite{zhang2019unsupervised} proposed an algorithm for 3D pose adaptation assuming the availability of depth and segmentation maps, Authors of \cite{jiang2021regressive} introduced RegDA for domain adaptive 2D pose estimation and \cite{kim2022unified} proposed an algorithm based on the Mean-Teacher framework \cite{tarvainen2017mean} (UDAPE) for the same, obtaining state-of-the-art results. Authors of \cite{raychaudhuri2023prior, peng2023source} proposed source-free algorithms for domain adaptive human pose estimation, achieving on par performance with UDAPE \cite{kim2022unified}. In this work, we use these algorithms to estimate pose keypoints on the unlabeled target dataset.

\noindent \textbf{UDA for Semantic Segmentation.} Unsupervised Domain Adaptation (UDA) has been extensively explored for urban scene semantic segmentation. One of the earliest pioneering works in this area was contributed by Vu {\it et. al.} \cite{vu2019advent}, who introduced the concepts of adversarial learning and entropy minimization to enhance domain adaptive segmentation in urban settings. Zou {\it et. al.} \cite{zou2019confidence} proposed the use of high-confidence pseudo-labels, while Zheng {\it et. al.} \cite{zheng2021rectifying} employed uncertainty guidance for self-training regularization. Kim {\it et. al.} \cite{kim2020learning} improved AdvEnt \cite{vu2019advent} by leveraging confident pseudo-labels and style-transfer, advancing the field. Recent works, DACS \cite{tranheden2021dacs} and SePiCo \cite{xie2023sepico}, achieved state-of-the-art results on established benchmarks. Source-free approaches, as suggested by Liu {\it et. al.} \cite{liu2021source} and Huang {\it et. al.} \cite{huang2021model}, have also gained attention, although these methods indirectly utilize source data. While domain adaptive semantic segmentation for urban scenes has advanced significantly, there remains a research gap in adapting segmentation for human body parts. We show that existing algorithms excel in urban scene segmentation but struggle with segmenting human body parts.

%% file: sections/updated_method.tex
\section{Method}
\label{sec:method}
In this section, we provide a detailed description of our proposed algorithm (\PIPS) for human body part segmentation under distribution shifts. Our primary aim is to learn a human parsing model $\mathcal{F}$, which takes an RGB image $x \in$ $\mathbb{R}^{H \times W \times 3}$ as input, and predicts the segmentation mask $y = \mathcal{F}(x)$, where $y \in$ $\mathbb{R}^{H \times W \times K}$ for each of the $K$ classes. Here, $H$ and $W$ refer to the spatial dimensions of the image and the resulting segmentation output. 

We assume access to a source dataset $\mathcal{S} = \{x_{i}, y_{i}, p_{i} \}_{i=1}^{M}$, where $x_{i}$ is an RGB image, $y_{i}$ and $p_{i}$ are the corresponding segmentation masks and pose keypoints, respectively. The pose keypoints provide additional anatomical information about the human in a given image. Note that, unlike the segmentation masks, the keypoints are \textit{sparse} annotations. Additionally, we are provided with an unlabeled target dataset $\mathcal{T}$, which contains images $ \{ x_{i} \}_{i=1}^{N}$ but lacks both body part segmentation masks and pose keypoint annotations. 

We seek to adapt a model that has been trained on the source dataset, denoted as $\mathcal{F_{\mathcal{S}}}$, to obtain improved performance on the target dataset $\mathcal{T}$, assuming the availability of source data. This adaptation process is driven by the anatomical information provided by the pose keypoints and is designed to yield superior results compared to directly applying $\mathcal{F_{\mathcal{S}}}$ on $\mathcal{T}$. Note that pose keypoints on $\mathcal{T}$ are \textit{not} available. This requires estimation of pose keypoints on $\mathcal{T}$ using a pre-trained pose estimation model $\mathcal{P}$ (Section~\ref{method-UDA-Pose}). We denote the model under adaptation for $\mathcal{T}$ as $\mathcal{F_{\mathcal{T}}}$. 

We also extend our algorithm (\PIPS) to the source-free setting \cite{liang2020we} and denote this algorithm as \PIPSSF~(Section~\ref{method-SFPIPS}). Here, we assume that only $\mathcal{F_{S}}$ is available and do not assume the availability the source dataset $\mathcal{S}$ for our extended algorithm \PIPSSF. \\

\begin{figure*}[t]
    \centering
    \includegraphics[width = 0.9\linewidth]{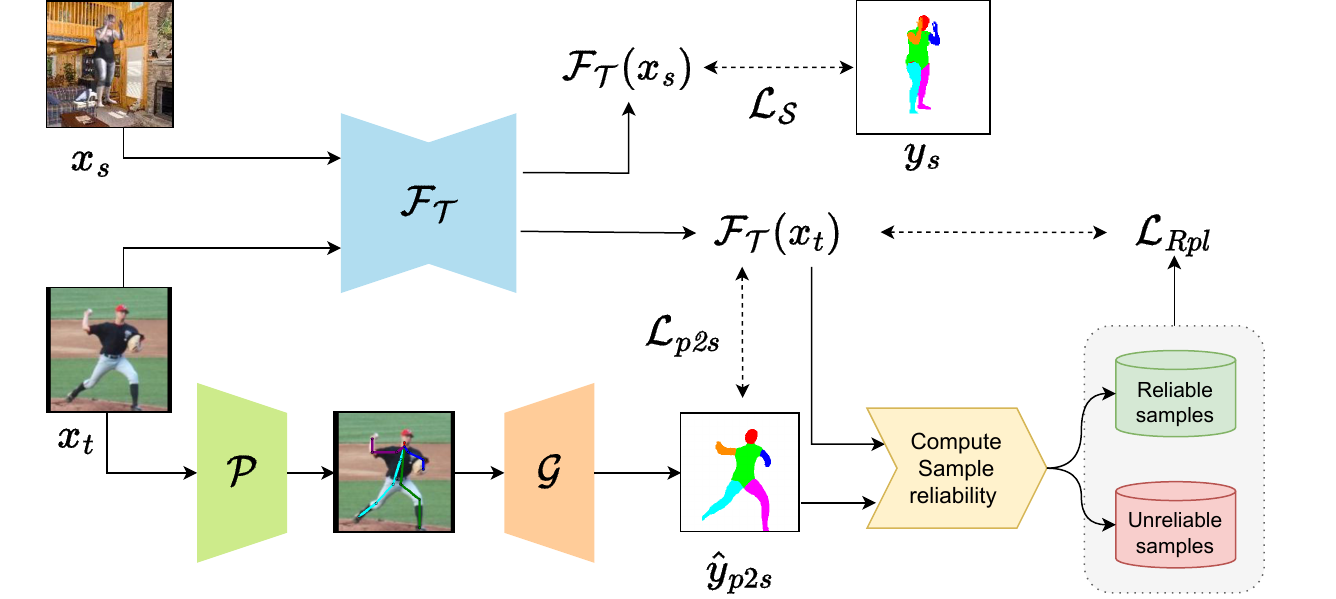}
    \caption{ \textbf{Overview of proposed framework.} Our proposed algorithm \PIPS~uses confident pseudo-labels from $\mathcal{F_{T}}$ to refine the predictions of the same on unlabeled data $\mathcal{T}$. Further, \PIPS~leverages anatomical context of the human body through pose estimates (obtained from $\mathcal{P}$) aligned with body part segmentation masks by $\mathcal{G}$. Note that \PIPS~can be extended to source-free settings, where $\{ x_{s}, y_{s} \}$ are absent (elaborated in Section~\ref{method-SFPIPS}). Even in the absence of source-data, the anatomical context obtained from $\mathcal{G} \circ \mathcal{P}$ acts as a strong regularizer, mitigating catastrophic forgetting and compensating for domain shifts.}
    \label{fig:method}
\end{figure*}

\noindent {\bf Overview:} We propose to use an anatomy guided pseudo-labelling technique to obtain improved human body part segmentation masks on $\mathcal{T}$. We note that, using the model $\mathcal{F_{\mathcal{S}}}$ directly on $\mathcal{T}$ results in notably inaccurate segmentation masks, as qualitatively illustrated in Figure~\ref{fig:teaser_qual}. Employing these inaccurate masks as pseudo-labels for adaptation would introduce error accumulation, ultimately leading to subpar performance on $\mathcal{T}$. Consequently, we introduce \PIPS, an enhanced pseudo-labeling technique that consists of two highly effective methods to circumvent this error accumulation issue: 
\begin{itemize} 

    \item In order to exploit the anatomical information of the human body, we propose to fuse information from the estimated pose-keypoints on $\mathcal{T}$ (see Section~\ref{method-UDA-Pose}). This allows $\mathcal{F_{\mathcal{T}}}$ to understand the human body parts better and eliminate inconsistencies caused by self-learning from confident yet incorrect pseudo-labels (see Section ~\ref{method-pose2seg}). 

    \item Further, we propose to use the human pose context for reliable pseudo-label selection, thereby disallowing $\mathcal{F_{\mathcal{T}}}$ to overfit to incorrect but confident pseudo-labels (see Section ~\ref{method-RPL}).

\end{itemize}
An overview of our method is presented in Figure~\ref{fig:method}.

\subsection{Pre-training source model}
We initialize the weights of $\mathcal{F}_{T}$ by the weights of a model that is trained on the labeled source domain in supervised fashion, referred to as the source model. This source model ($\mathcal{F}_{S}$) is trained by minimizing the loss described by Equation~\ref{eqn:sup}. 
\begin{equation}
    \mathcal{L}_{S} = \mathcal{L}_{ce}(\mathcal{F_{S}}(x_{s}), y) \ .
    \label{eqn:sup}
\end{equation}
Here, $\mathcal{L}_{ce}$ denotes the per-pixel cross-entropy loss.

\subsection{Self-learning from pseudo-labels}
\label{method-SL-PL}
Self-learning from pseudo-labels \cite{choi2019pseudo, wang2020unsupervised, zhang2021prototypical} offers a simple yet powerful solution in semi/self-supervised learning problems. The fundamental idea here is to utilize only highly confident pseudo-labels for learning as discussed in \cite{guillory2021predicting}. Pseudo-labels on $\mathcal{T}$, denoted by $\hat{y}_{pl}$ can be obtained by 
\begin{equation}
    \hat{y}_{pl}(\tau) = \argmax \mathbbm{1} [ \mathcal{F_{\mathcal{T}}}(x_{t}) \geq \tau ] \; \mathcal{F_{\mathcal{T}}}(x_{t}) \ ,
    \label{eqn:SL-PL}
\end{equation}  where $\tau$ is the confidence threshold and $\mathbbm{1}[.]$ is an indicator function. Note that the quality of the pseudo-labels is a function of the confidence threshold $\tau$.

These pseudo-labels are incorporated in the adaptation process by minimizing the loss $\mathcal{L}_{pl}$ defined by Equation~\ref{eqn:tgt-PL}, 
\begin{equation}
    \mathcal{L}_{pl}(\tau) = \mathcal{L}_{ce}(\mathcal{F_{\mathcal{T}}}(x_{t}), \hat{y}_{pl}(\tau)) \ .
    \label{eqn:tgt-PL}
\end{equation}
Nevertheless, we observe that depending solely on these pseudo-labels for the adaptation process results in sub-optimal performance, as it overlooks crucial anatomical information associated with the human body. Additionally, the incorporation of highly confident yet incorrect pseudo-labels can adversely affect the overall adaptation process.

\subsection{Anatomy guided self-training}
\label{method-pose2seg}

\begin{figure}[t]
    \centering
    \includegraphics[width=1\columnwidth]{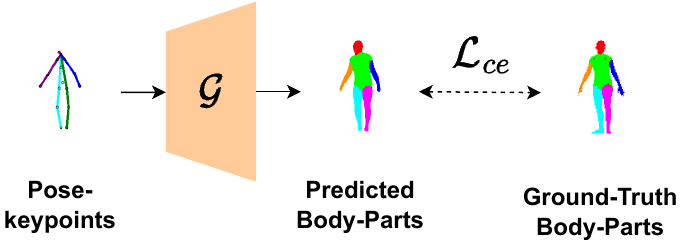}
    \caption{ \textbf{Learning algorithm for $\mathcal{G}$}. The model $\mathcal{G}$ is designed by predict segmentation maps $\hat{s} \in \mathbb{R}^{H \times W \times K}$ given pose keypoints $p \in \mathbb{R}^{P \times 2}$. $\mathcal{G}$ is trained using the cross-entropy loss with ground-truth segmentation maps $s \in \mathbb{R}^{H \times W \times K}$.}
    \label{fig:pose2seg}
    \vskip -4mm
\end{figure}

To utilize anatomical information from pose keypoints for adapting our model $\mathcal{F_{T}}$ to the target dataset $\mathcal{T}$, we propose learning a mapping $\mathcal{G}$ that transforms sparse pose keypoints to dense body part segmentation masks. Here, $\mathcal{G} : \mathbb{R}^{P \times 2} \shortrightarrow \mathbb{R}^{H \times W \times K}$, with $P$ representing the number of pose keypoints and $K$ denoting the number of classes in the segmentation maps (e.g., head, left/right leg, etc.).

The mapping $\mathcal{G}$ is learnt in a supervised fashion using our source dataset $\mathcal{S} = \{x_{i}, y_{i}, p_{i} \}_{i=1}^{M}$ by minimizing the loss in Equation~\ref{eqn:L-G}, 
\begin{equation}
        \mathcal{L_{G}} = \mathcal{L}_{ce}(\mathcal{G}(p_{j}), y_{j})  \ .
    \label{eqn:L-G}
\end{equation}
It is important to note that the learning of $\mathcal{G}$ does not use any RGB images, thus making it domain agnostic and hence, seamlessly applicable across different domains. Figure~\ref{fig:pose2seg} provides an overview for the learning algorithm of $\mathcal{G}$.

Using the learned mapping $\mathcal{G}$ we can extract anatomically plausible pseudo-labels on the target dataset $\mathcal{T}$, however, this requires access to pose keypoint annotations on $\mathcal{T}$. Since we do not have access to any annotations on $\mathcal{T}$, we propose to estimate the same using a pre-trained pose estimation model which has been adapted on $\mathcal{T}$ (see Section ~\ref{method-UDA-Pose} for details).


Mathematically, pseudo-labels $\hat{y}_{p2s}$ are obtained by \begin{equation}
    \hat{y}_{p2s} = \mathcal{G} \circ \mathcal{P} (x_{t}) \ ,
    \label{eqn:p2s}
\end{equation} and are used in the adaptation process by minimizing the loss function described in Equation~\ref{eqn:L-P2S}, 


\begin{equation}
    \mathcal{L}_{p2s} = \mathcal{L}_{ce}(\mathcal{F_{\mathcal{T}}}(x_{t}), \hat{y}_{p2s})
    \label{eqn:L-P2S} \ .
\end{equation} 

Incorporating the loss function from Equation~\ref{eqn:L-P2S} along with losses in Equations~\ref{eqn:sup} and~\ref{eqn:tgt-PL} contributes significantly to the improvement of $\mathcal{F_{T}}$ over the base source model. However, using pose information solely as a regularizer is insufficient, as $\mathcal{F_{T}}$ may still overfit to highly confident but incorrect pseudo-labels, leading to sub-optimal performance. Therefore, we recommend utilizing the anatomical context from Equation~\ref{eqn:p2s} to enhance the refinement of pseudo-labels $\hat{y}_{pl}$, thus driving additional performance enhancements.

\subsection{Pose-guided reliable pseudo-label selection}
\label{method-RPL}
In order to make our method more robust to incorrect pseudo-labels, we wish to employ an approach that helps $\mathcal{F_{T}}$ avoid learning from highly-confident incorrect pseudo-labels. We argue that the quality of pseudo-labels obtained using Equation~\ref{eqn:SL-PL} on a given batch of images from $\mathcal{T}$ is not homogeneous. In other words, for a batch of target images, a single threshold parameter $\tau$ may not be an optimal choice. Taking inspiration from Karim {\it et al.} \cite{karim2023c}, we argue that any batch of target images consists of reliable and unreliable samples. In other words, predictions for some samples are relatively more accurate as compared against other samples from the same batch. Thus, we propose to split a given batch of target images (denoted by $x_{t}^{\mathbf{B}}$, containing $\mathbf{B}$ images) into reliable and unreliable splits containing $\mathbf{B}^{r}$ and $\mathbf{B}^{u}$ images respectively.

These splits are obtained by comparing the mIoU \cite{rezatofighi2019generalized} of the predictions $\mathcal{F_{T}}(x_{t})$ with $\hat{y}_{p2s}$. Essentially, for the $j^{th}$ image in a batch of target images, we obtain a score $h^{j}$ defined by Equation~\ref{miou-com},
\begin{equation}
    \label{miou-com}
    h^{j} = mIoU(\mathcal{\argmax(F_{T}}(x^{j}_{t})), \hat{y}^{j}_{p2s}) \ .
\end{equation}
This score $h^{j}$ is then used to decompose a batch of $\mathbf{B}$ images into reliable and unreliable splits as follows, 
\begin{equation}
    x_{t}^{\mathbf{B}}= 
    \begin{dcases}
         x_{t}^{r}, & \text{$h_{j}$} \geq \gamma\\
         x_{t}^{u}, & \text{otherwise}
    \end{dcases} \ ,
    \label{eqn:split}
\end{equation}
where $\gamma$ is the threshold defining the boundary between reliable and unreliable samples. 

This is followed by extracting the confident pseudo-labels using Equation~\ref{eqn:SL-PL} with confidence thresholds $\alpha$ and $\beta$ with $\beta \geq \alpha$, for reliable and unreliable splits respectively. These pseudo-labels are now incorporated into the framework using the loss function defined below,
\begin{equation}
    \begin{split}
        \mathcal{L}_{Rpl} & = \lambda \mathcal{L}_{ce}(\mathcal{F_{T}}(x_{t}^{r}), \hat{y}_{pl}(\alpha)) + (1 - \lambda) \mathcal{L}_{ce}(\mathcal{F_{T}}(x_{t}^{u}),  \hat{y}_{pl}(\beta)) \ .
    \label{eqn:loss-RPL}
    \end{split}
\end{equation}
Here, $\lambda$ is set to $\frac{|\mathbf{B^{r}}|}{|\mathbf{B}|}$. We experimentally demonstrate that this strategy of pseudo-label selection during adaptation contributes significantly to the overall performance gains of \PIPS~.

\subsection{Estimating pose keypoints on target dataset}
\label{method-UDA-Pose}

As described in \cite{zhang2019unsupervised, jiang2021regressive, kim2022unified, raychaudhuri2023prior}, pose estimation models often face challenges related to domain shift. Given that we rely on pose keypoints derived from a pre-existing pose estimation model in the target domain $\mathcal{T}$ to extract anatomical information, adapting the pose model to $\mathcal{T}$ is preferable for enhancing performance. To this end, we use UDAPE \cite{kim2022unified} for \PIPS~to estimate pose keypoints on the target dataset $\mathcal{T}$. However, it is important to note that, \emph{while adaptation of the pose estimation model ($\mathcal{P}$) to the particular target domain ($\mathcal{T}$) gives superior human body part segmentation masks, this step is not mandatory to the success of \PIPS. In fact, in Section~\ref{sec:abl} (Table~\ref{tab:imp-poseadapt}) we show that an unadapted pose estimation model can also deliver significant performance gains.} 

\subsection{Overall Objective Function}
\label{sec-overallobj}
Our final adaptation objective for \PIPS~is given below,
\begin{equation}
    \min_{\mathcal{F_{T}}} \;\; \eta_{1} \mathcal{L}_{S} + \eta_{2} \mathcal{L}_{p2s} + \eta_{3} \mathcal{L}_{Rpl} \ ,
    \label{eqn:overall}
\end{equation}
where $\{\eta_{i}\}_{i=1}^3$ are hyperparameters which control the contribution of individual loss terms. 

\subsection{Extending \PIPS~to source-free settings}
\label{method-SFPIPS}

In this section, we explore the extension of \PIPS~to the source-free setting \cite{liang2020we}, denoted as \PIPSSF. Crucially, in the context of \PIPSSF, the absence of $\mathcal{S}$ precludes the utilization of $\mathcal{L_{S}}$ (Equation~\ref{eqn:sup}) in training the model $\mathcal{F_{T}}$ (see Equation~\ref{eqn:overall-SF}). Additionally, as outlined in Section~\ref{method-pose2seg}, the parametric prior ($\mathcal{G}$) is typically learned using the source dataset $\mathcal{S}$. Consequently, the lack of $\mathcal{S}$ would hinder the training of $\mathcal{G}$. However, it is crucial to note that $\mathcal{G}$ does not rely on RGB images during training and is inherently domain agnostic. Even in the absence of $\mathcal{S}$, $\mathcal{G}$ could be trained using a set of paired segmentation masks ($y$) and pose keypoints ($p$) from any auxiliary dataset, similar to \cite{raychaudhuri2023prior}. Note that, assuming access to paired segmentation masks ($y$) and pose keypoints ($p$) does not violate any privacy concerns and also, significantly reduces storage requirements compared to storing the entire source dataset. 

Note that, UDAPE \cite{kim2022unified} is not applicable for \PIPSSF~for pose estimation on the unlabeled target dataset ($\mathcal{T}$) because UDAPE inherently relies on using source data. Consequently, we employ POST \cite{raychaudhuri2023prior}, a recent source-free domain adaptive human pose estimation algorithm, to predict pose keypoints on $\mathcal{T}$.

For \PIPSSF, $\eta_{1}$ in Equation~\ref{eqn:overall} is set to 0, as the source dataset is absent in source-free settings. Thus, our objective for \PIPSSF~reduces to \begin{equation}
        \min_{\mathcal{F_{T}}} \;\; \eta_{2} \mathcal{L}_{p2s} + \eta_{3} \mathcal{L}_{Rpl} \ .
    \label{eqn:overall-SF}
\end{equation}

%% file: sections/updated_experiments.tex
\section{Experiments}
\label{sec:exp}

In this section, we demonstrate both \PIPS's~and \PIPSSF's~ impressive ability to perform human body part segmentation on unlabeled target datasets in the presence and absence of source data respectively. We provide qualitative and quantitative analysis on multiple domain adaptive scenarios, alongside an ablation study highlighting the importance of individual components of our framework.

\subsection{Datasets}

\begin{itemize}
    \item {\bf SURREAL} (SUR) \cite{varol2017learning} is a large scale synthetic dataset with over six million frames containing humans performing various activities in indoor settings with labels for depth, body parts, surface normal and 2D/3D pose keypoints. 

    \item {\bf Unite the People} (UP) \cite{Lassner:UP:2017} is a compilation of multiple real-world datasets such as MPII \cite{andriluka14cvpr} and more containing approximately 8000 images. The dataset poses a challenge due to a large number of images captured in uncontrolled, real-world conditions, resulting in significant variability in the dataset.

    \item {\bf Humans3.6M} (H36M) \cite{h36m_pami} is an extensive real-world video dataset, comprising more than 3 million frames of 11 professional actors engaged in various actions, including walking, eating, and more under controlled indoor settings. Similar to \cite{kim2022unified, raychaudhuri2023prior}, we employ subjects 'S1,' 'S5,' 'S6,' 'S7,' and 'S8' for training, while subjects 'S9' and 'S11' are reserved for testing. 

    \item {\bf Leeds Sports Pose Dataset} (LSP) \cite{johnson2010clustered} contains 2000 images of humans involved in different sporting activities with corresponding pose keypoint annotations. Corresponding body part annotations are provided by \cite{Lassner:UP:2017}. 

    \item {\bf BRIAR} \cite{cornett2023expanding} is a recent comprehensive biometric dataset containing human images captured outdoors with natural degradations like atmospheric turbulence. Qualitative results are provided on this dataset due to the absence of ground-truth segmentation masks. 
    
\end{itemize}

\subsection{Baselines} 

We compare our proposed algorithms (\PIPS~and \PIPSSF) against three state-of-the-art domain adaptive semantic segmentation algorithms: \textbf{AdvEnt} \cite{vu2019advent}, \textbf{DACS} \cite{tranheden2021dacs} and \textbf{SePiCo} \cite{xie2023sepico}. We also benchmark against two supervised human parsing algorithms, \textbf{JPPNet} \cite{liang2018look}, and \textbf{DeepLabv3} \cite{chen2017rethinking}, and a recently proposed weakly-supervised algorithm by \textbf{Zhao {\it et al.}} \cite{zhao2022pose}. We also report the result for one additional baseline: {\it Source only}, which represents the model's performance when trained only on the source data.

\subsection{Implementation Details}

\noindent {\bf Human body part segmentation model ($\mathcal{F_{S}}$ and $\mathcal{F_{T}}$).} We adopt DeepLabv3 \cite{chen2017rethinking} with ResNet101 backbone \cite{he2016deep} as the model architecture for both $\mathcal{F_{S}}$ and $\mathcal{F_{T}}$. For both \PIPS~and \PIPSSF, models are are trained for 30 epochs with the Adam optimizer \cite{kingma2014adam} with an initial learning rate of $1e-04$, decaying by a factor of $10$ after the $5^{th}$ and $20^{th}$ epochs. We use a batch-size of 32 with 100 iterations per epoch. Using grid search, we set $\tau = 0.8$ in equation~\ref{eqn:tgt-PL}, $\gamma = 0.25$ in Equation~\ref{eqn:split} and $\alpha = 0.75$, $\beta = 0.85$ in Equation~\ref{eqn:loss-RPL}. Also, $\eta_{i}$ in Equations~\ref{eqn:overall} and~\ref{eqn:overall-SF} is set at 1. All results are reported with the model from the last epoch. \\

\noindent {\bf Aligning pose keypoints and segmentation maps ($\mathcal{G}$).} $\mathcal{G}$ has a simple decoder architecture \cite{radford2015unsupervised} which employs a projection and reshape operation followed with strided convolutional layers with input tensor [$b_s$ $\times$ $\mathbf{P}$ $\times$ 2 $\times$ 1], yielding an output tensor [$b_s$ $\times$ $\mathbf{K}$ $\times$ 256 $\times$ 256]. and is trained with the Adam optimizer \cite{kingma2014adam} for 200 epochs on the SURREAL dataset \cite{varol2017learning} with a batch-size of 32 and a learning rate of $1e-04$. \\

\noindent \textbf{Pose estimation model ($\mathcal{P}$)}: We use the same architecture for $\mathcal{P}$ as used in \cite{kim2022unified, raychaudhuri2023prior} for our pose estimation model. All hyperparameters and augmentations were used as is proposed in \cite{kim2022unified, raychaudhuri2023prior} for this work. \\

\noindent \textbf{Environment}: We use the PyTorch coding environment for our algorithms, with all experiments being performed on a single Nvidia RTX-3090 GPU.

\noindent



\subsection{Quantitative Results} 

We evaluate \PIPS~and \PIPSSF~against baseline algorithms on four different domain adaptation scenarios and report their respective results in Tables~\ref{exp:SHU, exp:S2U, exp:H2U, exp:LSP}. We report the segmentation performance of all algorithms as an average mIoU \cite{rezatofighi2019generalized} across seven semantic classes: Background (BG), Head (HD), Torso (TR), Left hand (LH), Right Hand (RH), Left Leg (LL) and Right Leg (RL). For all Tables, best results are in \textcolor{blue}{\bf blue} and the second best results are in \textcolor{red}{\bf red}. 

\begin{table*}[!htb]
\caption{Quantitative Results (mIoU \cite{rezatofighi2019generalized}) for SURREAL \cite{varol2017learning} $\rightarrow$ Humans3.6M \cite{h36m_pami}.}
\centering
\begin{adjustbox}{max width=\linewidth} 
\begin{tabular}{@{}llcccccccc@{}}
\toprule
\multicolumn{1}{c}{\multirow{2}{*}{Algorithm}} & \multicolumn{1}{l}{\multirow{2}{*}{SF}} & \multicolumn{8}{c}{SURREAL $\rightarrow$ Humans3.6M} \\
\cmidrule(l){3-10} 
\multicolumn{1}{c}{} & \multicolumn{1}{l}{} & BG & HD & TR & LH & RH & LL & RL & Avg. \\
\midrule
{\it Source only} & - & 91.42 &	14.45 & 23.25 & 06.20 & 06.46 & 19.45 & 20.79 & 26.01 \\
{\it Oracle} & - & 98.40 & 60.04 & 73.57 & 49.05 & 50.68 & 66.64 & 70.46 & 66.98 \\
\midrule
AdvEnt \cite{vu2019advent} & \xmark & 96.47 & 44.30 & 60.51 & 29.63 & 32.04 & 42.02 & 43.1 & 49.72 \\
DACS \cite{tranheden2021dacs} & \xmark & 94.69 & \textcolor{red}{\bf 48.35} & 44.36 & 23.11 & 27.71 & 35.11 & 34.85 & 44.03 \\
SepiCo \cite{xie2023sepico} & \xmark & 95.98 & \textcolor{blue}{\bf 73.19} & 58.57 & 21.09 & 18.11 & 19.20 & 17.73 & 43.41 \\
\midrule
\PIPS & \xmark & \textcolor{blue}{\bf 97.19} & 44.29 & \textcolor{blue}{\bf 64.28} & \textcolor{blue}{\bf 41.22} & \textcolor{blue}{\bf 41.25} & \textcolor{blue}{\bf 52.38} & \textcolor{blue}{\bf 54.48} & \textcolor{blue}{\bf 56.44} \\
\PIPSSF & \cmark & \textcolor{red}{\bf 96.98} & 45.45 & \textcolor{red}{\bf 62.83} & \textcolor{red}{\bf 39.66} & \textcolor{red}{\bf 39.17} & \textcolor{red}{\bf 50.46} & \textcolor{red}{\bf 50.55} & \textcolor{red}{\bf 55.01} \\
\bottomrule
\end{tabular}
\end{adjustbox}
\label{exp:S2H}
\end{table*}

\begin{table*}[!htb]
\caption{Quantitative Results (mIoU \cite{rezatofighi2019generalized}) for SURREAL \cite{varol2017learning} $\rightarrow$ Unite the People \cite{Lassner:UP:2017}.}
\centering
\begin{adjustbox}{max width=\linewidth} 
\begin{tabular}{@{}llcccccccc@{}}
\toprule
\multicolumn{1}{c}{\multirow{2}{*}{Algorithm}} & \multicolumn{1}{l}{\multirow{2}{*}{SF}} & \multicolumn{8}{c}{SURREAL $\rightarrow$ UP} \\
\cmidrule(l){3-10} 
\multicolumn{1}{c}{} & \multicolumn{1}{l}{} & BG & HD & TR & LH & RH & LL & RL & Avg. \\
\midrule
{\it Source only} & - & 90.51 & 31.11 & 40.52 & 18.03 & 17.81 & 25.22 & 25.74 & 35.56 \\
\midrule
AdvEnt \cite{vu2019advent} & \xmark & 93.82 & 49.33 & 60.47 & 34.32 & 33.86 & 43.27 & 42.87 & 51.13 \\
DACS \cite{tranheden2021dacs} & \xmark & 93.85 & 53.28 & 61.74 & 33.40 & 34.45 & 37.99 & 38.94 & 50.60 \\
SepiCo \cite{xie2023sepico} & \xmark & 93.94 & \textcolor{red}{\bf 55.99} & 63.92 & 23.63 & 13.90 & 28.66 & 22.46 & 43.21 \\
\midrule
\PIPS & \xmark & \textcolor{blue}{\bf 95.25} & \textcolor{blue}{\bf 57.14} & \textcolor{blue}{\bf 70.16} & \textcolor{blue}{\bf 44.78} & \textcolor{blue}{\bf 43.89} & \textcolor{blue}{\bf 55.34} & \textcolor{blue}{\bf 53.86} & \textcolor{blue}{\bf 60.06} \\
\PIPSSF & \cmark & \textcolor{red}{\bf 94.98} & 52.09 & \textcolor{red}{\bf 68.90} & \textcolor{red}{\bf 42.02} & \textcolor{red}{\bf 44.46} & \textcolor{red}{\bf 54.28} & \textcolor{red}{\bf 52.89} & \textcolor{red}{\bf 58.52} \\
\bottomrule
\end{tabular}
\end{adjustbox}
\label{exp:S2U}
\end{table*}

\begin{table*}[!htb]
\caption{Quantitative Results (mIoU \cite{rezatofighi2019generalized}) for Humans3.6M \cite{h36m_pami} $\rightarrow$ Unite the People \cite{Lassner:UP:2017}}.
\centering
\begin{adjustbox}{max width=\linewidth} 
\begin{tabular}{@{}llcccccccc@{}}
\toprule
\multicolumn{1}{c}{\multirow{2}{*}{Algorithm}} & \multicolumn{1}{l}{\multirow{2}{*}{SF}} & \multicolumn{8}{c}{Humans3.6M $\rightarrow$ UP} \\
\cmidrule(l){3-10} 
\multicolumn{1}{c}{} & \multicolumn{1}{l}{} & BG & HD & TR & LH & RH & LL & RL & Avg. \\
\midrule
{\it Source only} & - & 51.95 & 05.77 & 11.20 & 02.54 & 01.64 & 04.37 & 04.58 & 11.72 \\
\midrule
AdvEnt \cite{vu2019advent} & \xmark & 93.49 & \textcolor{red}{\bf 54.69} & 60.19 & 29.63 & 30.33 & 35.09 & 33.74 & 48.17  \\
DACS \cite{tranheden2021dacs} & \xmark & 94.04 & 53.14 & 64.94 & 33.93 & 34.64 & 40.89 & 41.32 & 51.84  \\ 
SePiCo \cite{xie2023sepico} & \xmark & 92.84 & 52.46 & 62.54 & 21.03 & 15.26 & 18.24 & 21.35 & 40.53  \\ 
\midrule 
\PIPS & \xmark & \textcolor{red}{\bf 94.11} & \textcolor{blue}{\bf 54.72} & \textcolor{blue}{\bf 69.48} & \textcolor{blue}{\bf 42.47} & \textcolor{blue}{\bf 42.43} & \textcolor{red}{\bf 49.06} & \textcolor{red}{\bf 50.28} & \textcolor{blue}{\bf 57.51}  \\
\PIPSSF & \cmark & \textcolor{blue}{\bf 94.90} & 51.04 & \textcolor{red}{\bf 68.29} & \textcolor{red}{\bf 40.24} & \textcolor{red}{\bf 42.70} & \textcolor{blue}{\bf 53.63} & \textcolor{blue}{\bf 51.59} & \textcolor{red}{\bf 57.34}  \\ 
\bottomrule
\end{tabular}
\end{adjustbox}
\label{exp:H2U}
\end{table*}

\noindent In Tables~\ref{exp:S2H}, \ref{exp:S2U} and~\ref{exp:H2U}, we analyze three distinct domain adaptation scenarios: adapting from a synthetic dataset \cite{varol2017learning} to a real dataset captured in controlled indoor settings \cite{h36m_pami}, adapting from a synthetic dataset to a real dataset captured under uncontrolled settings \cite{Lassner:UP:2017}, and adapting from a real controlled dataset \cite{h36m_pami} to a real uncontrolled dataset \cite{Lassner:UP:2017}. In all these scenarios, the performance of the {\it Source only} model is notably insufficient, indicating the presence of a domain gap. \PIPS~consistently outperforms AdvEnt \cite{vu2019advent}, DACS \cite{tranheden2021dacs}, and SePiCo \cite{xie2023sepico}. Although all baseline algorithms significantly improve upon the {\it Source only} performance, they fall short of optimal results due to the absence of human anatomical context. In contrast, by incorporating anatomical information from pose keypoints, \PIPS~achieves optimal segmentation performance. \\

\begin{table*}[!htb]
\caption{Quantitative Results (mIoU \cite{rezatofighi2019generalized}) for unsupervised human body-part segmentation on the LSP \cite{johnson2010clustered} dataset. We use SURREAL \cite{varol2017learning} as our source dataset for this experiment. Here Sup. stands for supervision, $\mathbf{P}$ and $\mathbf{S}$ refer to supervision from ground-truth pose keypoints and segmentation masks respectively.}
\centering
\begin{adjustbox}{max width=\linewidth} 
\begin{tabular}{@{}llcccccccc@{}}
\toprule
\multicolumn{1}{c}{Algorithm} & \multicolumn{1}{c}{Sup.} & BG & HD & TR & LH & RH & LL & RL & Avg. \\
\midrule
JPPNet \cite{liang2018look} & $\mathbf{P}$ & \textcolor{red}{\bf 93.24} & \textcolor{red}{\bf 58.54} & 56.67 & 15.35 & 15.31 & 23.38 & 28.23 & 41.53 \\
DeepLabv3 \cite{chen2017rethinking} & $\mathbf{S}$ & 93.21 & \textcolor{blue}{\bf 63.02} & \textcolor{blue}{\bf 62.36} & 37.51 & 33.42 & 42.89 & 43.63 & 53.72 \\
Zhao {\it et. al.} \cite{zhao2022pose} & $\mathbf{P}$ & 92.59 & 54.57 & \textcolor{red}{\bf 58.56} & \textcolor{blue}{\bf 42.95} & \textcolor{red}{\bf 38.86} & \textcolor{red}{\bf 46.04} & \textcolor{red}{\bf 49.97} & \textcolor{red}{\bf 54.78} \\
\midrule 
\PIPS & \xmark & \textcolor{blue}{\bf 95.90} & 53.93 & 56.57 & \textcolor{red}{\bf 39.94} & \textcolor{blue}{\bf 40.52} & \textcolor{blue}{\bf 50.12} & \textcolor{blue}{\bf 50.05} & \textcolor{blue}{\bf 55.29} \\
\bottomrule
\end{tabular}
\end{adjustbox}
\label{exp:LSP}
\end{table*}

\noindent In Table~\ref{exp:LSP}, we evaluate \PIPS's body-part segmentation performance on the LSP dataset \cite{johnson2010clustered} and compare it to the baseline algorithms. Notably, \PIPS, operating in a self-supervised manner, achieves performance comparable to both weakly and fully supervised baselines. Moreover, methods such as \cite{zhao2022pose} require ground-truth pose keypoints for target data, potentially leading to sub-optimal performance without them. Also, obtaining pose keypoints for in-the-wild images is challenging and those errors would propagate into the segmentation algorithm \cite{zhao2022pose} \footnote{The code for \cite{zhao2022pose} is not publicly available, so we could not test its performance for estimated pose keypoints.}. Thus, this experiment underscores the effectiveness and scalability of \PIPS~in pragmatic settings. \\

\noindent{\textbf{Analysing \PIPSSF:}} In Tables~\ref{exp:S2H}, \ref{exp:S2U} and~\ref{exp:H2U}, \PIPSSF, the source-free variant of our proposed algorithm \PIPS, surpasses all existing baseline algorithms. This accomplishment is credited to the robust regularizing effect of anatomical information from pose keypoints, allowing \PIPSSF~to comprehend the underlying human body anatomy and avoid overfitting to highly confident yet anatomically incorrect pseudo-labels. The slight performance decrease in \PIPSSF~compared to \PIPS~is due to the lack of source data during adaptation. 

\begin{figure*}[!htb]
    \centering
    \includegraphics[width = 1\linewidth]{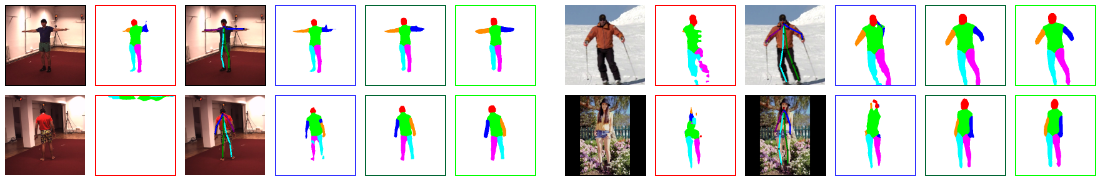}
    \caption{ \textbf{Qualitative Results on H36M\cite{h36m_pami} and UP\cite{Lassner:UP:2017} datasets.}: Left column show qualitative results for SUR \cite{varol2017learning} $\rightarrow$ H36M \cite{h36m_pami} and right column show qualitative results for SUR \cite{varol2017learning} $\rightarrow$ UP \cite{Lassner:UP:2017}. {\textit{Left to right}}: RGB image \cite{h36m_pami, Lassner:UP:2017}, {\it Source only} predictions, estimated pose keypoints using UDAPE \cite{kim2022unified} predictions of AdvEnt \cite{vu2019advent}, predictions from \PIPS, and predictions of \PIPSSF.}
    \label{fig:qual-1}
\end{figure*}

\begin{figure*}[!htb]
    \centering
    \includegraphics[width = 1\linewidth]{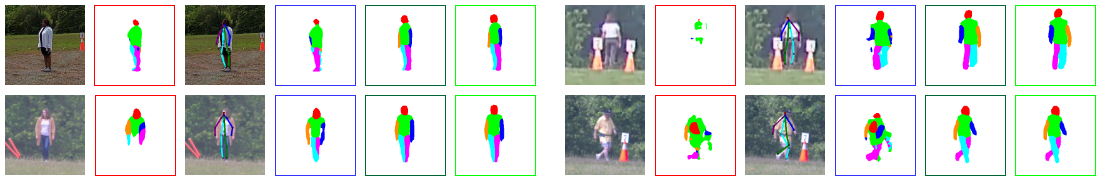}
    \caption{ \textbf{Qualitative Results on SUR \cite{varol2017learning} $\rightarrow$ BRIAR \cite{cornett2023expanding}.} {\textit{Left to right}}: RGB image \cite{cornett2023expanding}, {\it Source only} predictions, estimated pose keypoints using UDAPE \cite{kim2022unified} predictions of AdvEnt \cite{vu2019advent}, predictions from \PIPS, and predictions of \PIPSSF.}
    \label{fig:qual-2}
\end{figure*}

\subsection{Qualitative Results} 

Figure~\ref{fig:qual-1} displays qualitative outcomes for the domain adaptation scenarios SURREAL \cite{varol2017learning} $\rightarrow$ Humans3.6M \cite{h36m_pami} and SURREAL \cite{varol2017learning} $\rightarrow$ UP \cite{Lassner:UP:2017}. Figure~\ref{fig:qual-2} presents qualitative results for the SURREAL \cite{varol2017learning} $\rightarrow$ BRIAR \cite{cornett2023expanding} domain adaptation scenario. In both cases, it is evident that the performance of the {\it Source only} approach is glaringly inaccurate, thereby highlighting the problem of domain shifts between synthetic and real images. Observe that, AdvEnt \cite{vu2019advent} demonstrates notable enhancements compared to the {\it Source only} approach, showcasing its effectiveness in mitigating domain shifts through adversarial learning. Despite these notable improvements, it is important to observe that the predictions from AdvEnt \cite{vu2019advent} are anatomically implausible, as it does not consider the underlying anatomical structure of the human body. In contrast, by using pose keypoints as an anatomical prior, both \PIPS~and \PIPSSF, deliver optimal and anatomically plausible human body part segmentation masks. Please note that we exclusively present comparative qualitative results from AdvEnt \cite{vu2019advent} in Figures~\ref{fig:qual-1} and~\ref{fig:qual-2}, as it demonstrates the highest quantitative performance, surpassing other baseline algorithms. This facilitates comparison with the top-performing baseline algorithm.

\begin{table*}[!htb]
\caption{\textbf{Ablation study of individual loss terms.} Ablation study of individual loss terms used in this work on SURREAL \cite{varol2017learning} $\rightarrow$ Humans3.6M \cite{h36m_pami}. Note that $\mathcal{L}_{pl}$ is already included in $\mathcal{L}_{Rpl}$ and hence is not used explicitly in the overall framework.}
\centering
\begin{adjustbox}{max width=\linewidth}
\begin{tabular}{cccccccccccc}
\toprule
 $\mathcal{L_S}$ & $\mathcal{L}_{pl}$ & $\mathcal{L}_{p2s}$ & $\mathcal{L}_{Rpl}$ & BG & HD & TR & LH & RH & LL & RL & Avg. \\ \midrule
 \cmark & \xmark & \xmark & \xmark & 91.42 &	14.45 & 23.25 & 06.20 & 06.46 & 19.45 & 20.79 & 26.01        \\
 \cmark & \cmark & \xmark & \xmark & 95.71 & \textcolor{red}{\bf 42.75} & 54.97 & 22.29 & 23.56 & 37.80 & 38.12 & 45.03       \\
 \xmark & \xmark & \cmark & \xmark &  96.06 & 34.34 & 62.18 & 35.06 & 33.44 & 47.23 & 50.92 & 51.32    \\
 \cmark & \cmark & \cmark & \xmark & \textcolor{red}{\bf 96.23} & 33.98 & \textcolor{red}{\bf 63.11} & \textcolor{red}{\bf 36.31} & \textcolor{red}{\bf 35.10} & \textcolor{red}{\bf 48.95} & \textcolor{red}{\bf 52.26} & \textcolor{red}{\bf 52.28}    \\
 \cmark & -- & \cmark & \cmark &  \textcolor{blue}{\bf 97.19} & \textcolor{blue}{\bf 44.29} & \textcolor{blue}{\bf 64.28} & \textcolor{blue}{\bf 41.22} & \textcolor{blue}{\bf 41.25} & \textcolor{blue}{\bf 52.38} & \textcolor{blue}{\bf 54.48} & \textcolor{blue}{\bf 56.44}    \\
 \bottomrule
\end{tabular}
\end{adjustbox}
\label{tab:ablation}
\end{table*}

\subsection{Ablation Study}
\label{sec:abl}

\noindent{\textbf{Contribution of individual loss terms:}} In Table~\ref{tab:ablation}, we present an ablation study to analyse the significance of our proposed loss terms on the SURREAL \cite{varol2017learning} $\rightarrow$ Humans3.6M \cite{h36m_pami} benchmark. Note that, trivially adding the target pseudo-labels in the loss function (Equation~\ref{eqn:tgt-PL}) significant improves ($\approx$ 19 \%) the predictions on the target domain. Further, incorporating the anatomical context from the pose keypoints by using Equation~\ref{eqn:L-P2S}, leads to another 6 \% improvement. Observe that, anatomical information obtained from individual pose keypoints standalone delivers significant gains, underscoring its crucial role in the framework. Finally, our novel reliable pseudo labelling algorithm (Section~\ref{method-RPL}) improves the performance by another 4\%, thus highlighting its contribution to the overall framework. In summary, this study validates the contribution of every component of our framework in providing state-of-the-art performance gains for \PIPS. \\

\begin{table}
\centering
\caption{\textbf{Analysing the role of Pose Adaptation.} We analyze the importance of adapting pose models to the particular target domain for optimal results for SUR \cite{varol2017learning} $\rightarrow$ UP \cite{Lassner:UP:2017}.}
\label{tab:imp-poseadapt}
\begin{tabular}{cc}
\toprule
Algorithm & Avg. mIoU \\ \midrule
{\it Source only} & 35.56 \\
AdvEnt \cite{vu2019advent} & 51.13 \\
\PIPS~w/o Pose Adapt & \textcolor{red}{\bf 54.20} \\ 
\PIPS~& \textcolor{blue}{\bf 60.06} \\
\bottomrule
\end{tabular}
\end{table}

\noindent{\textbf{Analysing the role of pose adaptation:}} In Table~\ref{tab:imp-poseadapt}, we analyse the role of adapting the pose estimation model to the particular target domain under consideration (Section~\ref{method-UDA-Pose}). For both \PIPS~and \PIPSSF, pose keypoints provide the anatomical guidance which is crucial for our framework. Observe that, even the incorporation of an unadapted pose estimation model (trained exclusively on synthetic data \cite{varol2017learning}) within the \PIPS~framework provides an $\approx$ 3\% improvement over AdvEnt \cite{vu2019advent}, the best performing baseline algorithm. This shows that even a sub-optimal pose estimation model ($\mathcal{P}$), trained only on synthetic source data and not adapted to the specific target domain, can still offer reasonable pose estimates on unlabeled target data, thereby aiding \PIPS~in predicting reasonably accurate body part segmentation masks. \\ 

\begin{table}
\centering
\setlength{\tabcolsep}{6pt}
\begin{adjustbox}{max width=\linewidth}
\begin{tabular}{c|ccccc}
\toprule
$\alpha$, $\beta$ & 0.95,0.95 & 0.85,0.95 & \textbf{0.75,0.85} & 0.70,0.85 & 0.65,0.75 \\ 
\midrule
mIoU & 59.73 & 59.91 & \textbf{60.06} & 59.94 & 59.62 \\
\bottomrule 
\toprule
$\gamma$ & 0.10 & 0.20 & \textbf{0.25} & 0.30 & 0.40 \\
\midrule
mIoU & 59.62 & 59.81 & \textbf{60.06} & 59.93 & 59.97 \\
\bottomrule
\end{tabular}
\end{adjustbox}
\caption{\textbf{Sensitivity analysis for hyperparameters.} Performance variations in \PIPS~due to changes in hyperparameters for SUR \cite{varol2017learning} $\rightarrow$ UP \cite{Lassner:UP:2017}.}
\label{tab:hypr}
\end{table}

\noindent{\textbf{Hyperparameters:}} Table~\ref{tab:hypr} depicts a study of sensitivity analysis of the performance \PIPS~as a function of the hyperparameters $\alpha$, $\beta$ and $\gamma$ on the SURREAL \cite{varol2017learning} $\rightarrow$ UP \cite{Lassner:UP:2017} adaptation scenario. We note that \PIPS~is reasonably robust to changes in hyperparameters.

%% file: sections/5_conclusion.tex
\section{Conclusion}
\label{sec:concl}

We present \PIPS, an innovative domain adaptation algorithm for human body part segmentation in the presence of domain shifts. \PIPS~ leverages anatomical information from pose keypoints with uncertainty based pseudo-labelling to mitigate domain shifts. Unlike the state-of-the-art algorithms for semantic human parsing, \PIPS~ eliminates the need for expensive per-pixel annotations. Extensive experiments across multiple datasets shows \PIPS~ outperforms existing UDA algorithms, establishing it as a superior algorithm for human body part segmentation in the presence of domain shifts. We extend \PIPS~ to source-free settings (\PIPSSF), thereby ameliorating privacy and computational concerns associated with \PIPS, all while achieving comparable performance to \PIPS~. In conclusion, \PIPS~ and \PIPSSF~ present valuable self-supervised algorithms for human body part segmentation under domain shifts. \\
